\documentclass[letterpaper, 10 pt, conference]{ieeeconf}  % Comment this line out if you need a4paper

\IEEEoverridecommandlockouts                              % This command is only needed if 
\usepackage{graphics} % for pdf, bitmapped graphics files
\usepackage{amsmath} % assumes amsmath package installed
\usepackage{amssymb}  % assumes amsmath package installed
\usepackage{algorithm,algorithmic}
\usepackage{graphicx}
\usepackage{hyperref}
\usepackage{subfig}
\usepackage{xcolor}

\title{\LARGE \bf
A Generalized Robotic Handwriting Learning System based on Dynamic Movement Primitives (DMPs)
}

\author{Qian Luo$^{*,1}$, Jing Wu$^{*,1}$ and Matthew Gombolay$^{2}$% <-this % stops a space
\thanks{*The first two authors contributed equally}% <-this % stops a space
\thanks{$^{1}$Qian Luo is with the School of Electrical and Computer Engineering,
        Georgia Institute of Technology, North Ave NW, GA, US
        {\tt\small luoqian@gatech.edu}}%
\thanks{$^{1}$Jing Wu is with the School of Computer Science, College of Computing department,
        Georgia Institute of Technology, North Ave NW, GA, US
        {\tt\small jingwu@gatech.edu}}%
\thanks{$^{2}$Matthew Gombolay is faculty with the School of Interactive Computing, College of Computing          department, Georgia Institute of Technology, North Ave NW, GA, US
        {\tt\small matthew.gombolay@cc.gatech.edu}}%
}

\begin{document}

\maketitle
\thispagestyle{empty}
\pagestyle{empty}

%%%%%%%%%%%%%%%%%%%%%%%%%%%%%%%%%%%%%%%%%%%%%%%%%%%%%%%%%%%%%%%%%%%%%%%%%%%%%%%%
\begin{abstract}

Learning from demonstration (LfD) is a powerful learning method to enable a robot to infer how to perform a task given one or more human demonstrations of the desired task. By learning from end-user demonstration rather than requiring that a domain expert manually programming each skill, robots can more readily be applied to a wider range of real-world applications. Writing robots, as one application of LfD, has become a challenging research topic due to the complexity of human handwriting trajectories. In this paper, we introduce a generalized handwriting-learning system for a physical robot to learn from examples of humans' handwriting to draw alphanumeric characters. Our robotic system is able to rewrite letters imitating the way human demonstrators write and create new letters in a similar writing style. For this system, we develop an augmented dynamic movement primitive (DMP) algorithm, DMP*, which strengthens the robustness and generalization ability of our robotic system.

\end{abstract}

%%%%%%%%%%%%%%%%%%%%%%%%%%%%%%%%%%%%%%%%%%%%%%%%%%%%%%%%%%%%%%%%%%%%%%%%%%%%%%%%
\section{INTRODUCTION}
\label{sec:int}

% \cite{childrenteach} children teach hri\\
% \cite{buildingsuc}child-robot interactions learning by teaching\\
% \cite{johal2016child}\\
% \cite{rohrbeck2003peer}\\
% \cite{yin2016synthesizing} Inverse optimal control\\
% \cite{potkonjak2012robot} why handwriting robot?\\
% \cite{khansari2011learning} GMM
% \cite{williams2006extracting} HMM
% \cite{kulvicius2011modified} DMP refer

Handwriting-learning robot has become a significant research topic recent years, due to its wild range of usage in various areas \cite{childrenteach}\cite{potkonjak2012robot}\cite{buildingsuc}\cite{johal2016child}. Specifically, robots with the ability to imitate human handwriting trajectories are mostly used in child handwriting education \cite{childrenteach}\cite{rohrbeck2003peer}, where robots serve as both teachers and learners \cite{yin2016synthesizing}. In such 'learning from teaching' \cite{rohrbeck2003peer} scene, the robots could firstly teach children a handwriting criterion and then imitate children's demonstrations, which in return better helps the children to learn to write. In other fields of application, these robots are of vital significance in handwriting rehabilitation process, which requires exact finger-hand-arm coordination. Also, such robots could serve as a perfect test bed for handwriting bio-mechanic study \cite{potkonjak2012robot}.

The state-of-art robotic handwriting learning system is composed of movement capture system, learning algorithm and physical robots. The movement capture system could record whole trajectories of human demonstrations, which are then input to the learning algorithm. Then, physical handwriting robots implement the learned trajectories to write. Undoubtedly, to get better accuracy and the efficiency of the handwriting robots, powerful learning algorithm is required. However, learning human handwriting trajectories is quite a challenging task. For instance, the method we use to collect data should be accurate so as not to mislead the learning results, and more importantly, the algorithm we use should be adaptable to arbitrary shape and scale of the trajectories.

Many methods have been applied to deal with the robotic handwriting-learning problem. Gaussian Mixture Models (GMM) \cite{khansari2011learning} can describe the handwriting as a nonlinear time-invariant dynamical system and generalize over different start and end points of the trajectory. Williams et al. \cite{williams2006extracting} used Hidden Markov Models (HMM) to learn locally consistent demonstrations and extended it to multiple modes of writing a same letter. Inverse optimal control (IOC) \cite{yin2016synthesizing}\cite{2012levine} extracts implicit cost representations which are transferable for robots to develop control commands in various contexts.   
%\cite{zhao2020deep} including image processing algorithm and deep neural networks serve as effective solutions to this problem. However, these methods simply learn handwriting trajectories based on the while requiring massive demonstrations and computation resources. 

\begin{figure}[t]
\centering
\includegraphics[width=0.475\textwidth]{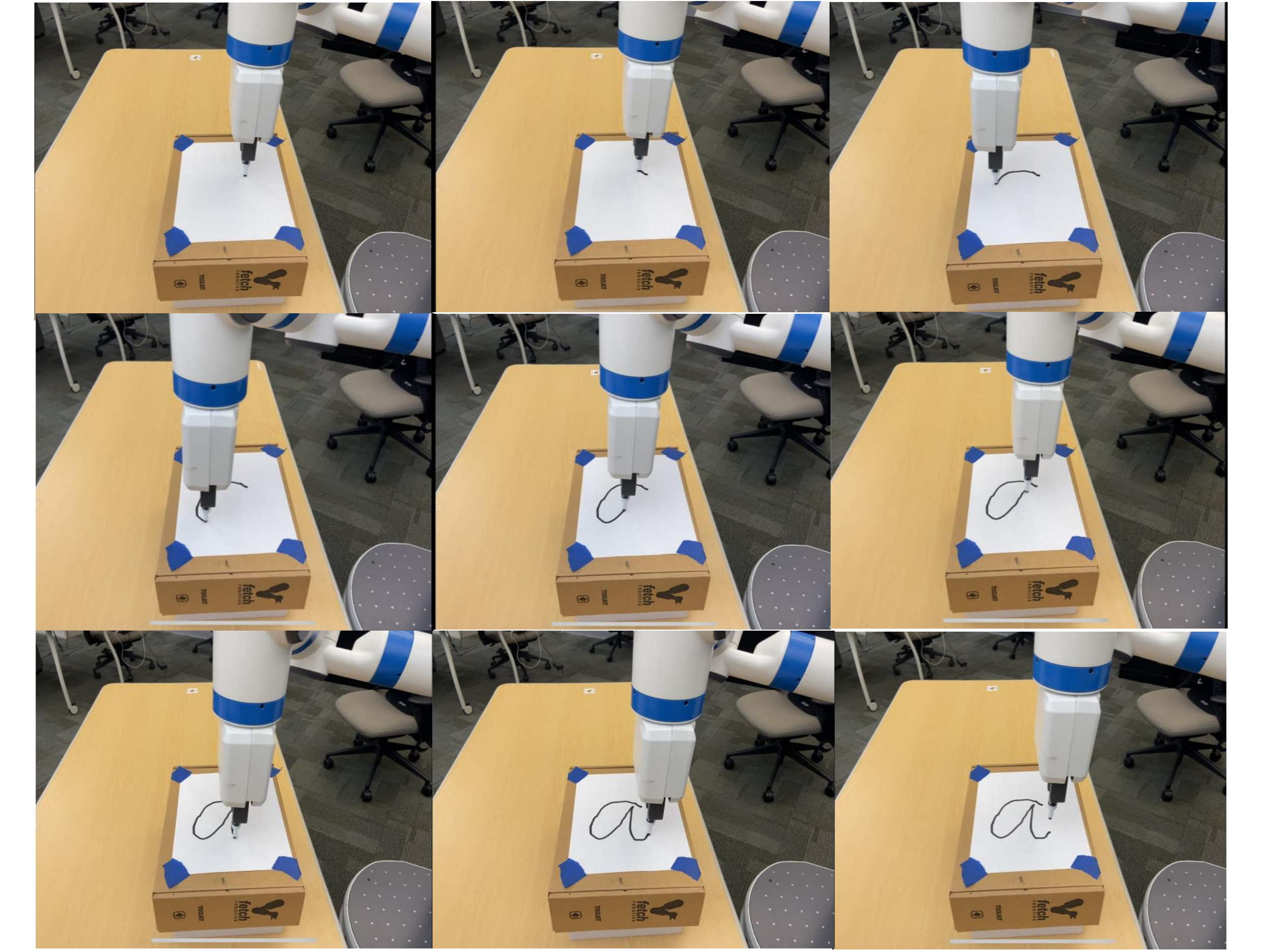}
\caption{This figure depicts our robot drawing the letter 'a'. Time progresses from left to right and then top to bottom.}\label{fig:robot_writing}
\end{figure}

Comparing to the methods mentioned above, Dynamic Movement Primitives (DMPs) \cite{ijspeert2002learning} serves as a more universal way in handwriting-learning problem for its flexibility and high efficiency in trajectory imitation problems\cite{kulvicius2011modified}. A standard DMPs uses a spring-damper system combined with a forcing function to be learned to match the movement trajectory. Previous work \cite{6008670} has shown that the use of such spring-damper system could enhance the stability and robustness of the generated motion. 

Hence, in this paper, we build our learning system based on DMPs algorithm to learn and reconstruct the movement trajectories of the whole handwriting process. To overcome some drawbacks of DMPs \cite{ginesi2019dmp++} and generalization ability and accuracy, we propose a modified version of DMPs: DMP*. DMP* is based on the optimal solutions in our study on the influence of Gaussian kernel shape and number. The implementation of DMP* algorithm is available at \url{https://github.com/jingwuOUO/dmp_writing}

There are three main contributions in our research.
\begin{enumerate}
\item We study how kernel shape and number influence learning performance and present a generalized, better way to solve the handwriting learning problem. 
\item We apply our algorithm to learn segmented handwriting strokes and create new letters of similar writing style. 
\item From collecting human writing data to utilizing real robot to reproduce the trajectory, our system forms a completed chain for robot to learn from human and write in real world, which is a meaningful application.
\end{enumerate}

We have introduced robotic handwriting-learning problem and the main ideas of our system in section \ref{sec:int} . Section \ref{sec:pre} show the preliminaries of DMP algorithm and related work. In section \ref{sec:method} we discuss our modifications of original formalism for DMPs algorithm and how these changes contribute to the improved performance. In section \ref{sec:evaluate} we introduce the architecture of our hardware system and show our experiment results of creating new letters based on our new DMPs algorithm. Section \ref{sec:conclusions} is our final conclusion 
% and section \ref{sec:futurework} is discussion about the future work.

\section{Preliminaries}
\label{sec:pre}

\subsection{Dynamic Movement Primitives}
% DMPs are units of actions that describe a particular movement trajectory and are formalized as stable attractor systems.
DMPs are a particular kind of dynamical formulation that can biologically imitate movement primitives of arbitrary shape in natural world by combining a spring-damper system and a forcing function together. DMPs has been widely applied in learning from demonstration problems, like table tennis \cite{muelling2010learning} \cite{kober2011reinforcement}, object grasping \cite{ude2010task}, and handwriting \cite{6008670}. Based on Ijspeert's \cite{ijspeert2002learning} and Schaal's \cite{SCHAAL2007425} previous work, the formulation of DMP is presented in Equations~\ref{cal_acc}-\ref{kerneleq}.
\begin{align} \ddot{y}& =\alpha_{z}(\beta_{z}(g-y)-\dot{y})+(g-y_{0})f \label{cal_acc} \end{align} 
\begin{align} f& = \frac{\sum\limits_{i=1}^{N}\Psi_{x}w_{i}}{\sum\limits^{N}_{i=1}\Psi_{x}}x  \label{weight} \end{align}
\begin{align} \dot{x}& =-\alpha_{x}x \label{decay_type} \end{align} 
\begin{align} \Psi_{x}& =\exp(-h_{i}(x-c_{i})^{2}) \label{kerneleq} \end{align}

In Equation~\ref{cal_acc}, $\alpha_{z}(\beta_{z}(g-y)-\dot{y})$ denotes the spring-damper system, where $\alpha_{z}$, $\beta_{z}$ are positive constants, $g$ is goal position, $y$ and $\dot{y}$ are current position and velocity. $(g-y_{0})f$ is the non-linear function used to fit the acceleration $\ddot{y}$, where $y_{0}$ is the initial position. 
% $\tau$ the temporal scaling factor, 
Equations~\ref{cal_acc}-\ref{kerneleq} show the composition of non-linear function $f$, which consists of weighted average of Gaussian kernels and a exponential decay factor $x$. In these equations, $\alpha_{x}$ is positive constant, $h_{i} \in \{h_{i} | i \in \{1,2,\ldots,N \} \}$ denotes the shape of Gaussian  kernel $\Psi_{i} \in \{ \Psi_{i} | i \in \{1,2,\ldots,N\} \}$, and $\omega_{i} \in \{ \omega_i | i \in \{1,2,\ldots,N\} \}$ are the learnable weights, which are determined by Locally Weighted Regression (LWR) to minimize the overall error:
\begin{align} \omega_i = \frac{x^T \Psi_{x} f}{x^T \Psi_{x} x} \label{lwr}\end{align}

According to {\bf Algorithm\ref{original_DMP}}, we first calculate all the $x$ in Equation~\ref{decay_type} and $f$ the non-linear part based on the input state $y$, $\dot{y}$ and $\ddot{y}$. Then we can get all the $\omega$ learned from the demonstration input by Locally Weighted Regression. At last we can use the Euler's equation to deduce all the $y$, $\dot{y}$ and $\ddot{y}$ with start state (start velocity and position), end position given and $\omega$ learned from previous processes.
\begin{algorithm}[t!]
    \caption{ Original DMP Algorithm  }
        \begin{algorithmic}[1]
        \STATE {\bf Input: }$\Theta = [\theta_t, \Dot{\theta_t}, \Ddot{\theta_t}], t \in \{ 1, \dots, T_{final}\}$\\
        \STATE Set $\alpha_z$ and $\beta_z$ \\
        % \FOR{all degree of freedoms $d$}
            \FOR{each parameter $\omega_i$}
                \STATE Set all the constants $h_i$ and $c_i$ \\
                \STATE Set $g=\theta_{final}$ and $y_0 = \theta_0$ \\
                % \STATE Calculate $y$, $\Dot{y}$ and $\Ddot{y}$ \\
                \STATE Calculate $x_t$  by integrating $\dot{x} =-\alpha_{x}x $ for all $t$\\
                \STATE Calculate $f$ by $\ddot{y} =\alpha_{z}(\beta_{z}(g-y)-\dot{y})+f$ for all $t$ \\
                \STATE Calculate $\Psi_{x} =\exp(-h_{i}(x-c_{i})^{2})$ for all $t$ \\
                \STATE Calculate $\omega_i$ by $\omega_i = \frac{x^T \Psi_{x} f}{x^T \Psi_{x} x}$
            \ENDFOR
        % \ENDFOR
        \STATE {\bf return} $\ddot{y_t} =\alpha_{z}(\beta_{z}(g-y_{t-1})-\dot{y}_{t-1})+f$, $\dot{y}_t = \dot{y}_{t-1} + \ddot{y}_{t-1} dt$ and $y_t = y_{t-1} + \dot{y}_{t-1}dt$ for all $t$ \\
        \end{algorithmic}
\label{original_DMP}
\end{algorithm}
.

\begin{figure*}[t]
% \centering
     \subfloat[\label{subfig-1}]{%
       \includegraphics[width=0.25\textwidth]{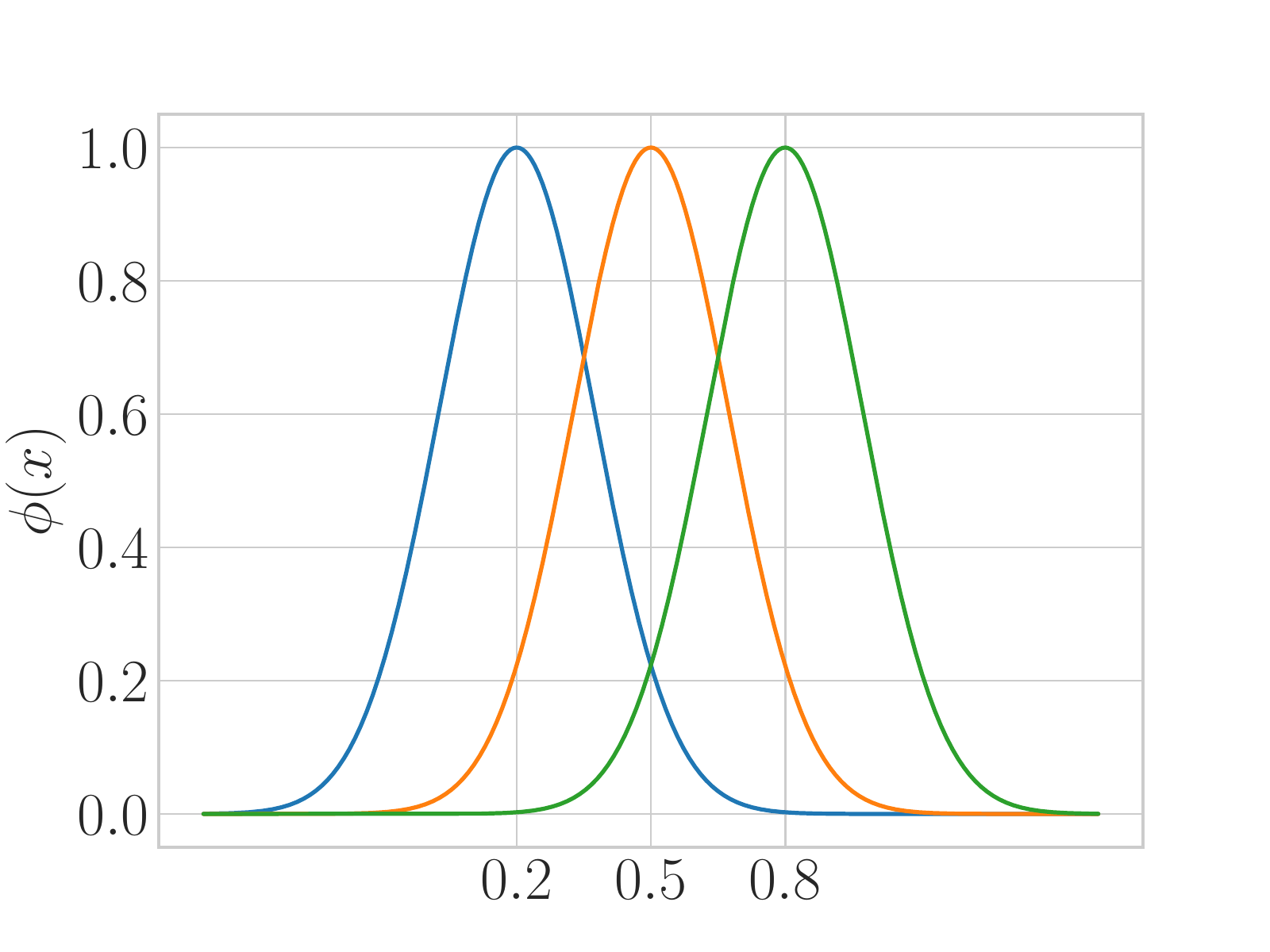}
     }
     \subfloat[\label{subfig-2}]{%
       \includegraphics[width=0.25\textwidth]{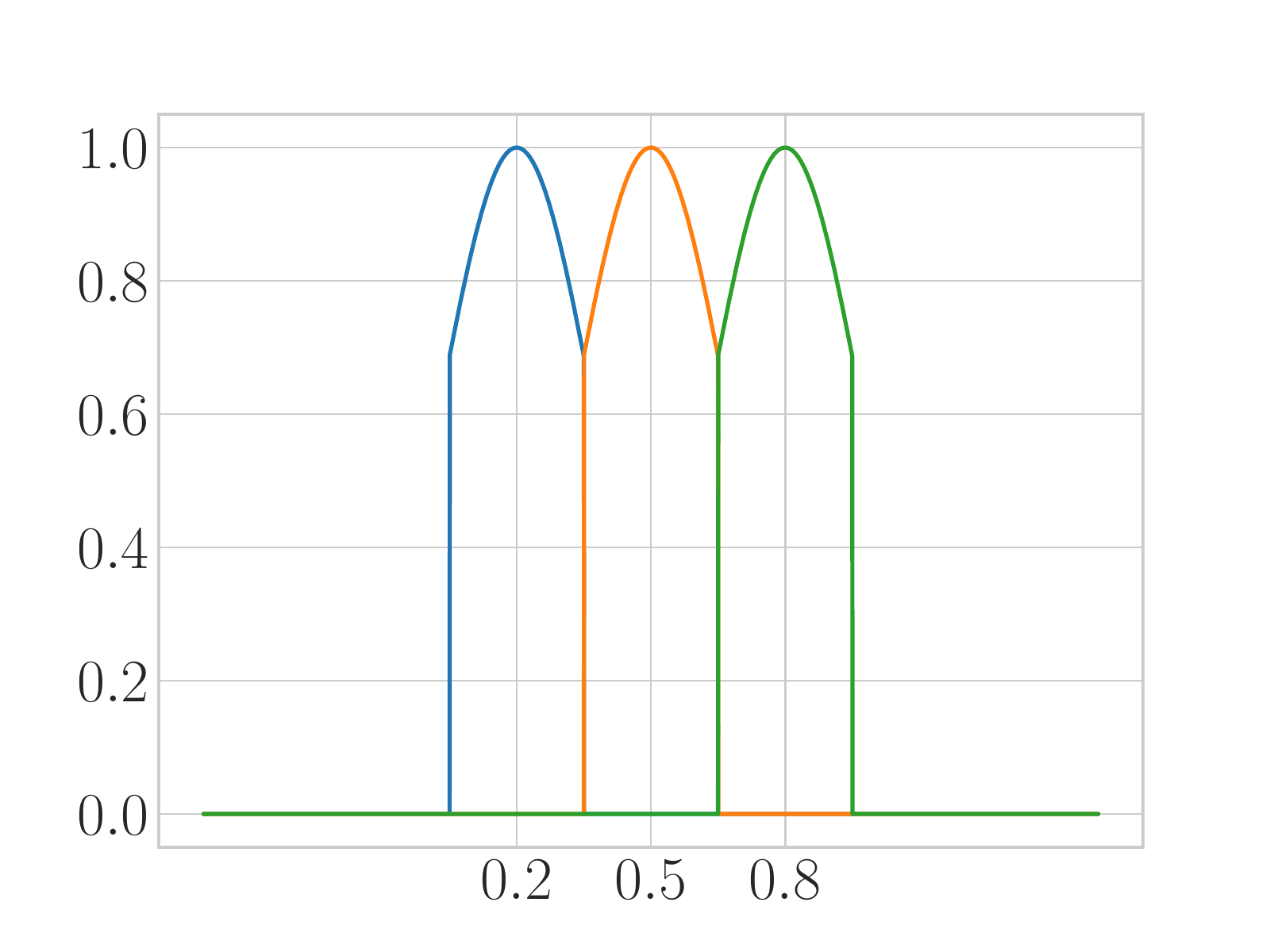}
     }
     \subfloat[\label{subfig-3}]{%
       \includegraphics[width=0.25\textwidth]{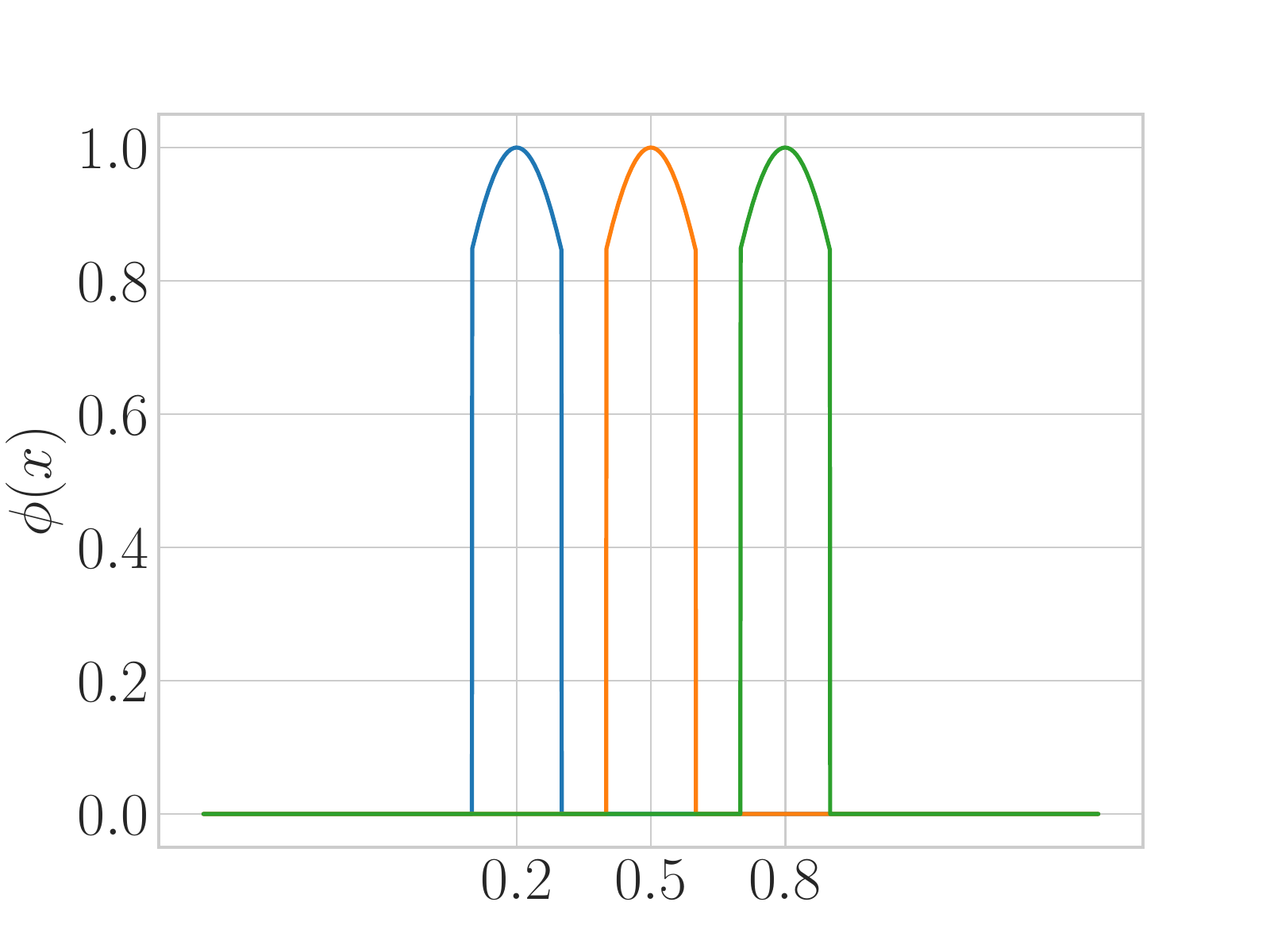}
     }
     \subfloat[\label{subfig-4}]{%
       \includegraphics[width=0.25\textwidth]{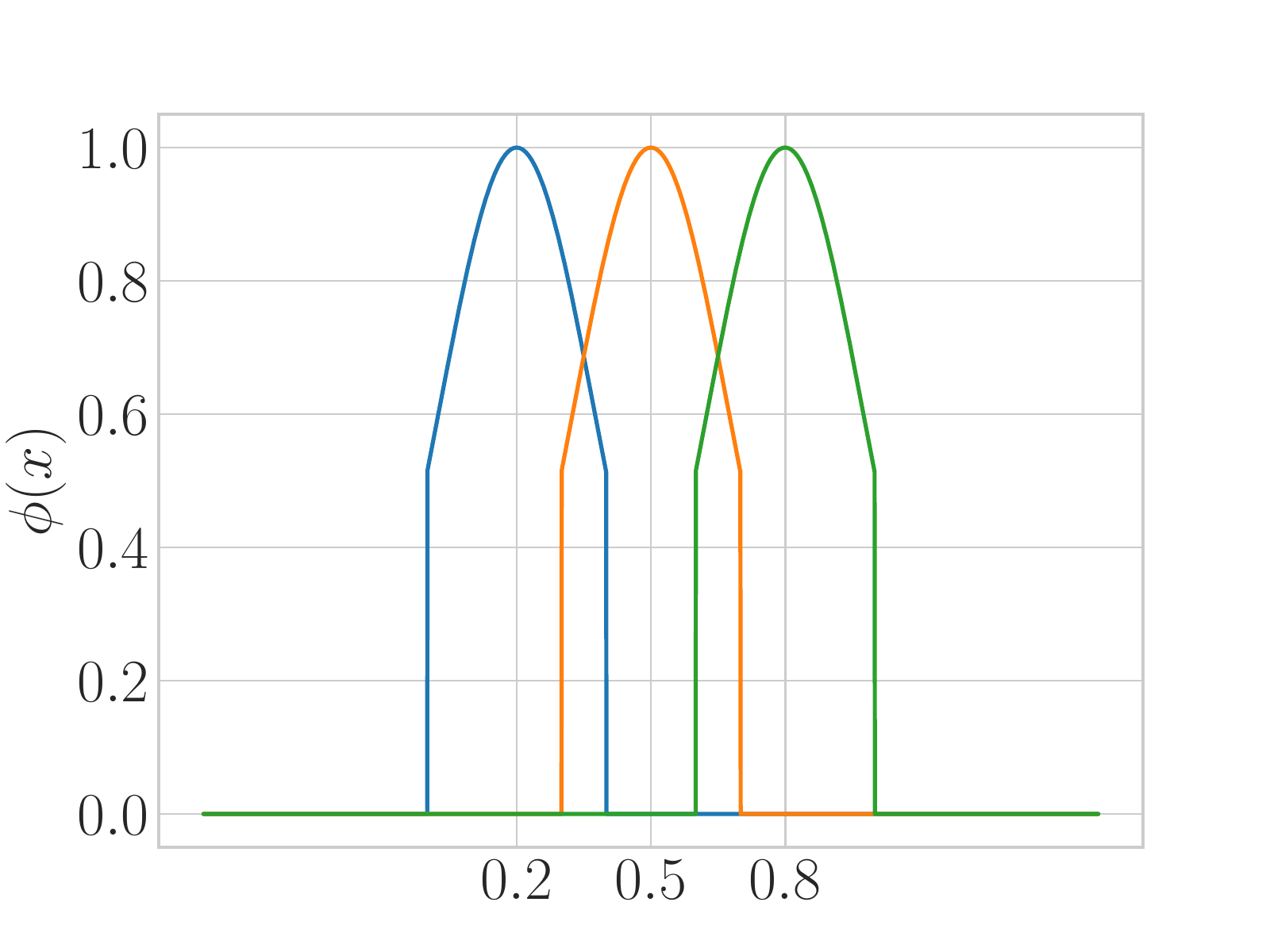}
     }
\caption{This figure depicts different kernel types for the DMP forcing function $f$. From left to right, the four kinds of kernel represent Full Gaussian kernel in (a), truncated Gaussian Kernel without overlapping but covers the full space in (b), truncated Gaussian Kernel without full space covered in (c), truncated Gaussian Kernel with overlapping part in (d).}\label{fig:kernel_intro}
\vspace{.5\baselineskip}
\end{figure*}

\subsection{Modifications to Original DMPs}
In previous studies \cite{kulvicius2011modified} \cite{ginesi2019dmp++}, simply applying the original formulations of DMPs to trajectory learning problem will not achieve good performance because of the complexity of the trajectory and the uncertainty of the relative position of start point and end point. To achieve better performance, many research papers present novel modifications on the original formulation of DMPs. Kulvicius et al.~\cite{6008670} used a joining version of DMPs to segment the handwriting letters and create the separate parts of letters. They also applied sigmoidal decay function instead of exponential decay function in the forcing function. Wang et al.~\cite{7759554} used a truncated version of DMPs (cutting both sides of Gaussian kernel) to achieve single letter imitation. Based on such truncated model, Wu et al.~\cite{Wu2018MultiModalRA} used linear decay system in the forcing function to reduce the kernels needed to calculate the trajectory.

Although the methods could replicate the demonstrated trajectories  in some cases, there is significant room to improve based on these prior approaches: First, these methods only proposed their models with certain fixed hyperparameters, like Gaussian kernel numbers and the width of each kernel, without study further about the most suitable choices of such hyperparameters. Secondly, in cases when the position of desired ending point are quite different from demonstrations \cite{ginesi2019dmp++}, these method may fail to create trajectories of similar shape, which is a basic requirement in learning from demonstration problems. To solve these two problems, we firstly study further into the influence of kernel shape and number based on previous research \cite{Wu2018MultiModalRA}. Then we try to enhance the robustness of the algorithm and propose a new formulation, DMP*, algorithm to overcome the drawbacks of previous methods. 

\begin{algorithm}[b!]
    \caption{ DMP* Algorithm  }
        \begin{algorithmic}[1]
        \STATE{{\bf Input: }$\Theta_d = [\theta_t, \Dot{\theta_t}, \Ddot{\theta_t}], t \in \{ 1, \dots, T_{final}\}, d \in \{x, y, z\}$}
        \STATE{Set kernel number $i = \frac{N}{10}$, $\alpha_z$ and $\beta_z$}
        \FOR{all degree of freedoms $d$}
            \FOR{each parameter $\omega_i$}
            \STATE{Set all the constants $h_i$ and $c_i$}
            \STATE{Set $g=\theta_{final}$ and $y_0 = \theta_0$}
            % \STATE Calculate $y$, $\Dot{y}$ and $\Ddot{y}$ \\
            \STATE{Calculate $x_t$  by integrating $\dot{x} =-\alpha_{x}x $ for all $t$}
            \STATE{Calculate $f$ by $\ddot{y} =\alpha_{z}(\beta_{z}(g-y)-\dot{y})-\alpha_{z}\beta_{z}(g-y_{0})x+\alpha_{z}\beta_{z}f$ for all $t$}
            \STATE{ Calculate $\Psi_{x}$ by Eq.(\ref{kernel_shape}) for all $t$}
            \STATE Calculate $\omega_i$ by $\omega_i = \frac{x^T \Psi_{x} f}{x^T \Psi_{x} x}$ \\
            \ENDFOR
        \ENDFOR
        \STATE {\bf return} $\ddot{y_t} =\alpha_{z}(\beta_{z}(g-y_{t-1})-\dot{y}_{t-1})-\alpha_{z}\beta_{z}(g-y_{0})x+\alpha_{z}\beta_{z}f$, $\dot{y}_t = \dot{y_{t-1}} + \ddot{y}_{t-1} dt$ and $y_t = y_{t-1} + \dot{y}_{t-1}dt$ for all $t$ \\
        \end{algorithmic}
\label{dmp*_DMP}
\end{algorithm}

\section{METHOD}
\label{sec:method}

\begin{figure}[t]
\centering
\includegraphics[width=0.5\textwidth]{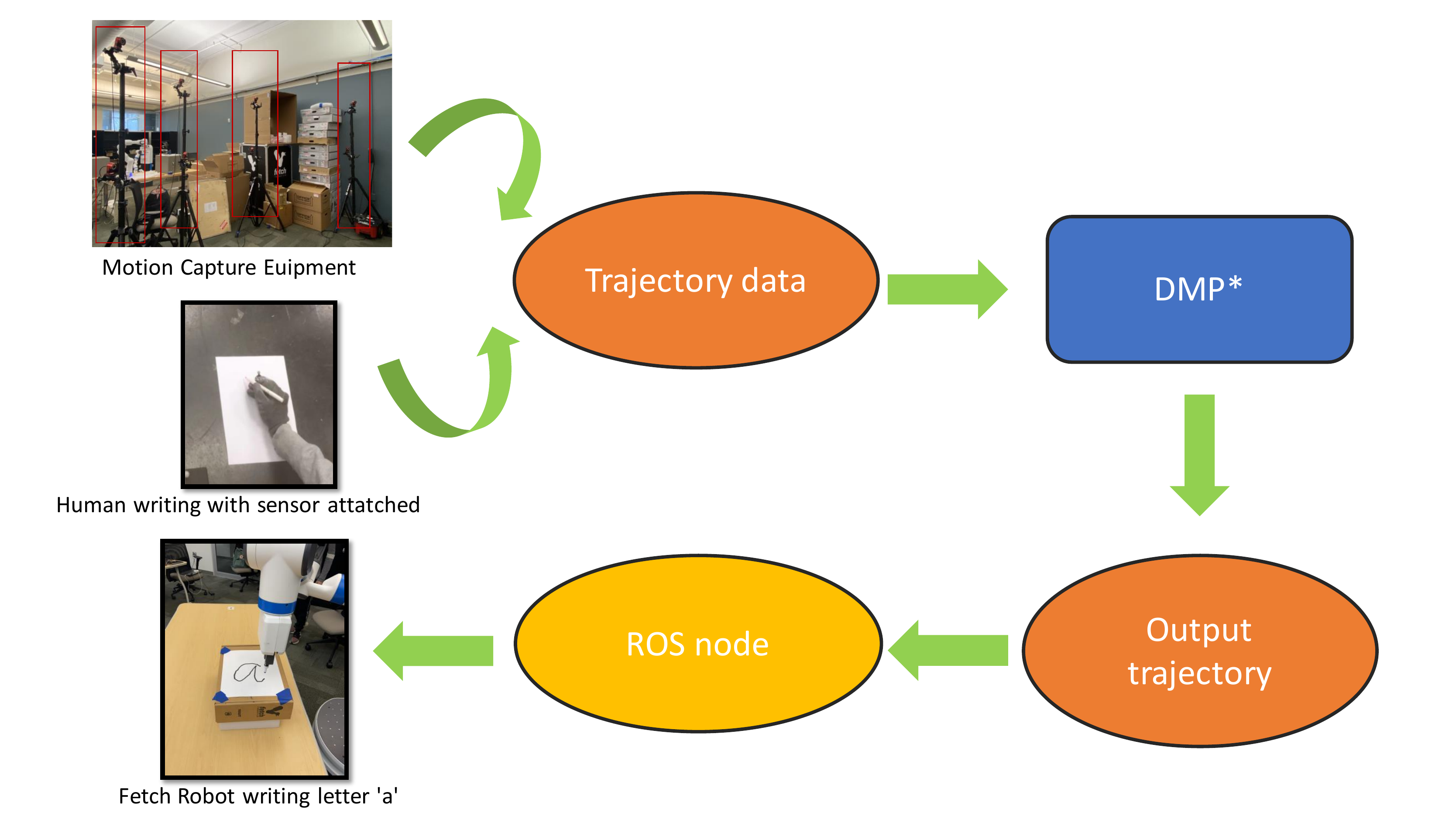}
\caption{This figure depicts the architecture of our robotic handwriting learning system.}\label{fig:pipeline}
\end{figure}

In this section, we describe the changes we make to the original DMP algorithm in formulating our improved, DMP*, method. In particular, we describe how and why we augment the DMP kernel shape and quantity (Section \ref{sec:kernelShapeNum}). Next, we describe how we enable DMP* to be robust to changing the goal position for the motion (Section \ref{sec:GoalChanging}). Finally, we show how we can naturally provide stroke segmentation to build up an inventory of primitive characters (e.g., 'I' and 'D') to compose more complex characters (e.g., 'P') in Section \ref{sec:strokeSeg}. Pseudocode for DMP* is provided in Algorithm \ref{dmp*_DMP}. An overview of our training pipeline is shown in Figure \ref{fig:pipeline}, which we describe further in Section \ref{sec:evaluate}.

\subsection{Changing the Kernel Shape and Number}
\label{sec:kernelShapeNum}
In this section, we further study the influence of different types of truncated kernel shape \cite{Wu2018MultiModalRA} and the suitable number of kernels to learn a trajectory of certain sample points. We will leverage the insights we glean from this kernel exploration to propose a generalized formulation of DMPs to learn handwriting trajectories.   

\paragraph{The influence of kernel shape}
Typically, truncated Gaussian kernels \cite{7759554}\cite{Wu2018MultiModalRA}, could be described as:
\begin{align}\Psi_{i}& =\begin{cases} \exp(-\frac{h_{i}}{2} (x-c_{i})^{2}),& \text{if}\ x-c_{i}\leq\theta_{i}\\ 0, & \text{otherwise} \end{cases} \label{kernel_shape} \end{align}
where $c_{i}$ $h_{i}$ $\theta_{i}$ are fixed based on a certain kernel number\cite{7759554}. Though this type of kernel has been proved to be able to enhance the accuracy of DMPs in \cite{7759554}, the research is lack of sufficient analysis of the influence of kernel width, i.e.~the shape of truncated kernels. 

In this section, we will study further about this point. As is shown in Figure~\ref{fig:kernel_intro}, there are three types of kernel width setting compared with original Gaussian kernels Figure~\ref{subfig-1}. In Figure~\ref{subfig-2} the adjacent kernels are concatenated without overlapping. The second case is shown in Figure~\ref{subfig-3} where the adjacent kernels are not concatenated. The third case is that the kernels are concatenated and adjacent kernels are overlapped, as is shown in Figure~\ref{subfig-4}.

In this study, we set the centers of each Gaussian kernels $c_{i}$ to be evenly distributed in $[0,1]$ and the total kernel number be $N$. Obviously, when the kernel width is ${1/N}$ (${\theta_{i}=1/2N}$), the adjacent kernels are concatenated without overlapping, which is the case in Figure~\ref{subfig-2}.

To figure out how kernel width will influence the learning performance, we study the learned trajectories of 5 letters: 'a', 'B', 'D', 'e', 'M' based on Gaussian kernels of different width. In this study, we apply linear decay\cite{Wu2018MultiModalRA} to DMPs time decay mechanism rather than exponential decay system: 
\begin{align}\dot{x}& =-x/T \end{align}

\begin{figure}[t!]
\centering
\includegraphics[width=0.475\textwidth, height = 7cm]{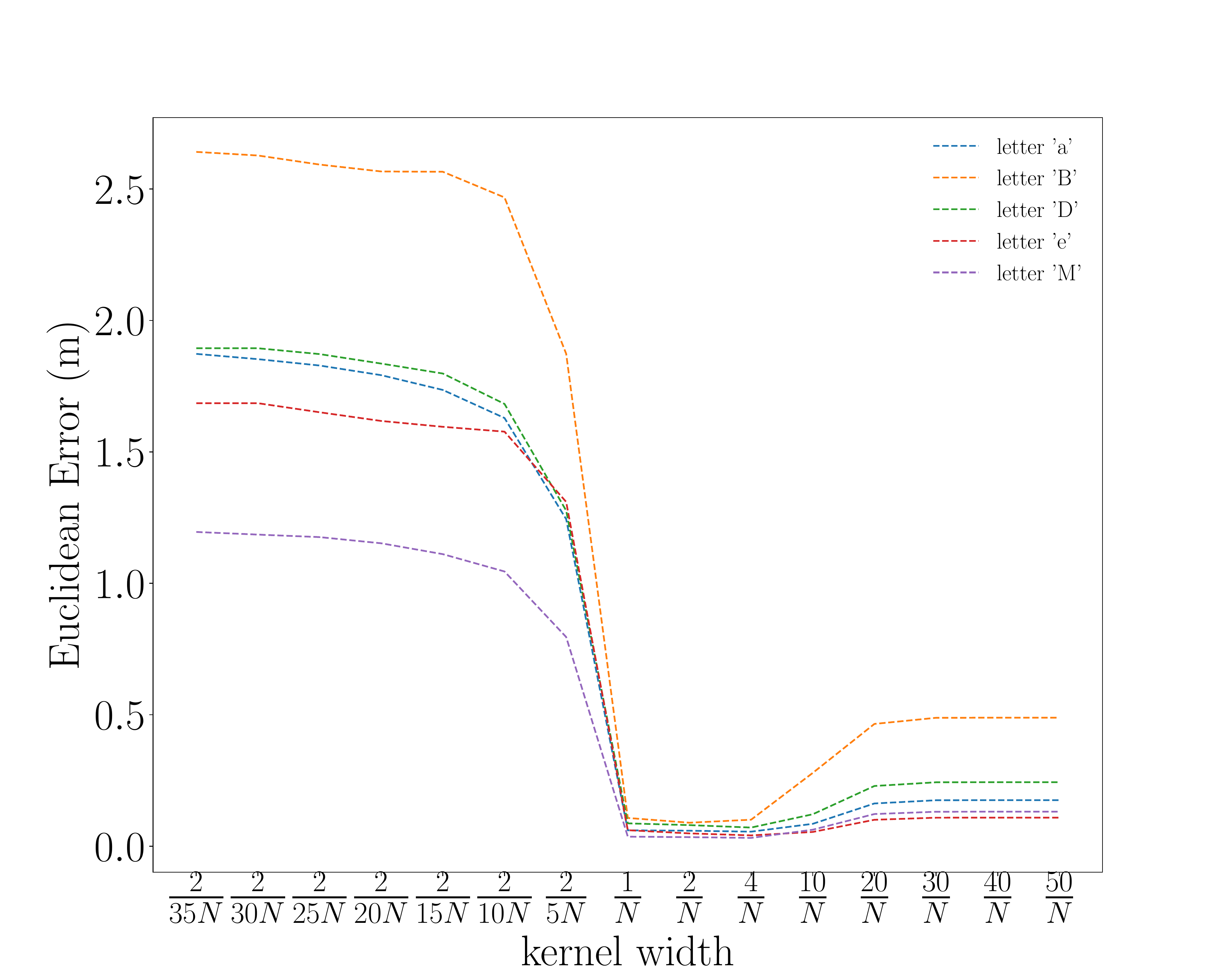}
\caption{This figure depicts the euclidean distance (i.e.,~error) between the desired and actual trajectories when varying kernel width. Different colors denote different letters. In all cases, the accuracy is the highest when the kernel width is between $1/N$ and $4/N$. }\label{fig:kernel_width}
\end{figure}

Figure~\ref{fig:kernel_width} shows how the euclidean error changes according to the kernel width. For all the 5 letters, when the kernel width is smaller than $1/N$, the euclidean error is large but decays with the increase of kernel width. When the kernel width is between $1/N$ and $4/N$, the error reaches the minimum. And the error increases when the kernel width is larger than $4/N$, and finally, when the kernel width reaches $20/N$, the error becomes stable.

\begin{figure}[t!]
\centering
\includegraphics[width=0.475\textwidth, height=7cm]{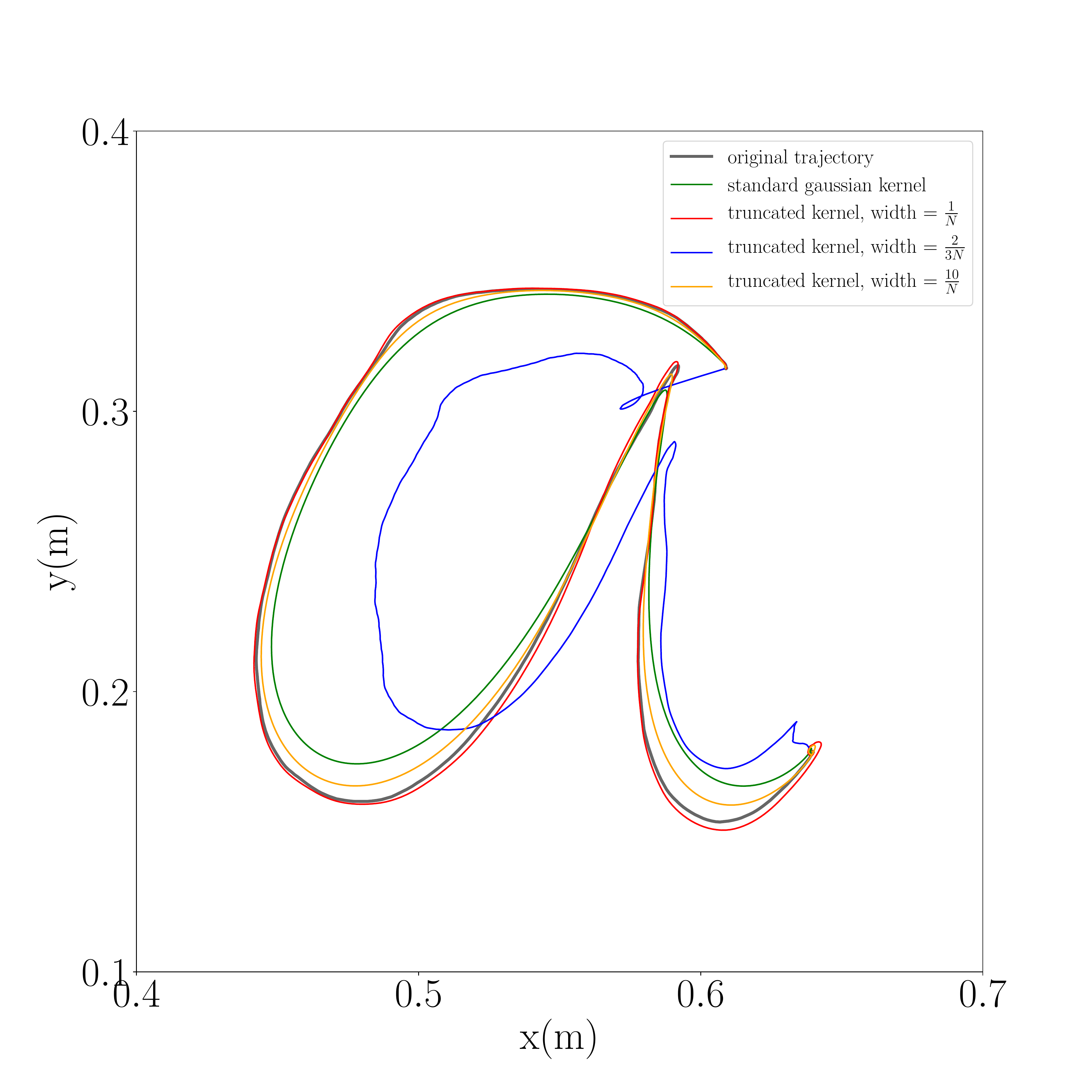}
\caption{This figure depicts the imitation performance of writing letter 'a' varying the kernel width. Truncated kernel with $1/N$ kernel width shows the best result.}\label{fig:kernel_shape}
\end{figure}

To show the imitation performance of different kernel width(${1/N}$ ${2/3N}$ ${10/N}$) against the conventional kernel baseline, we take letter 'a' studied in Figure~\ref{fig:kernel_width} as an example. As shown in Figure~\ref{fig:kernel_shape}, we find that DMPs perform best with ${1/N}$ kernel width and when kernel width equals to ${10/N}$, DMPs perform worse than former case but better than the original Gaussian formulation. The DMPs perform the worst among the four when kernel width equals to ${2/3N}$. 

% \vspace*{0.1em}

We infer from above that the truncated DMPs performs the best when the kernel width is between $1/N$ and $4/N$, not too narrow or too wide. The main drawback of simply applying Gaussian kernel is that the width of each kernel is too large which will result in the inference between multiple kernels. In other words, for a given state $\langle y_t,\dot{y}_t \rangle $ from a trajectory for training a DMP, the learned behavior for that state is influenced by multiple kernels, which means each kernel is involved in the optimization of several states. Thus it is hard to calculate the weights of each kernel to minimize overall loss. Thus, using a truncated formulation of Gaussian kernel will make each kernel more 'independent' and less impacted by adjacent kernels, enabling simpler and more feasible optimization of overall weights. When kernels are separated (i.e., not concatenated to adjacent kernels), the performance is much worse than other three cases, since there are many places where $\Psi_{i}=0$, which leads to 0 weights in these places. However, the use of our method is limited to use in learning relatively little data, due to the bias introduced by less complex truncated models. So we have to increase the kernel number to eliminate the bias. In the next part, we will find out the least number of kernels for certain sample points to get satisfying performance.

\paragraph{The influence of the number of kernels, $N$}
\begin{figure}[t!]
\centering
\includegraphics[width=0.475\textwidth, height = 7cm]{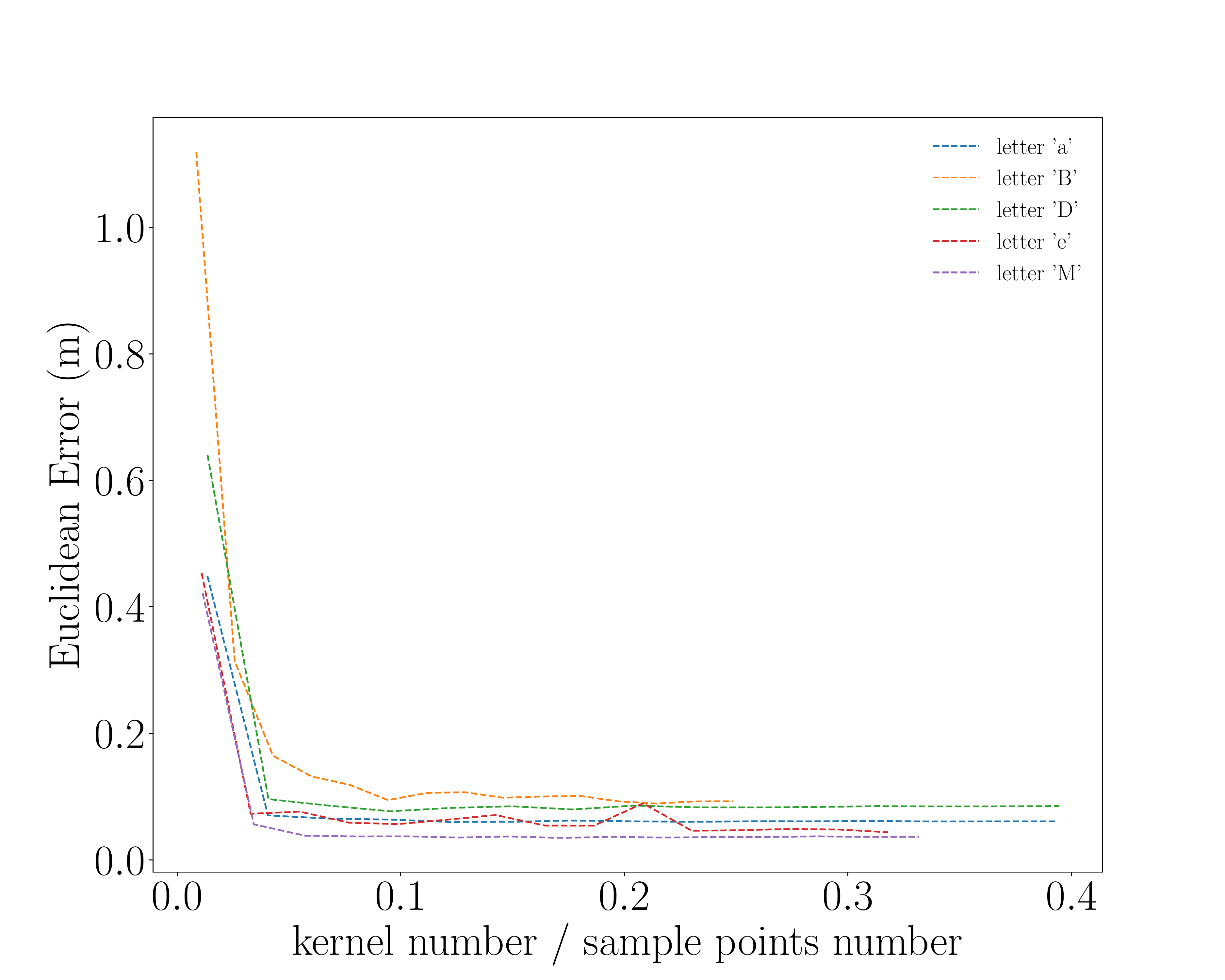}
\caption{This figure depicts the change in euclidean error when kernel number increases. The error reaches a stable, low level when kernel number reaches $\frac{1}{10}^{th}$ the number of total sample points.}\label{fig:kernel_number}
\end{figure}

In previous work \cite{6008670}, researchers acknowledged that the larger the kernel number $N$, the more subtle and accurate the imitating trajectory could be. However, few research studied the influence of larger range of kernel numbers and generalized the connection between the number of sample points in demonstrations and kernel number to acquire relatively good performance. 

To address this gap in prior literature, we explore the influence kernel numbers have on euclidean errors. Firstly we fix the kernel width to ${1/N}$, then we apply DMPs of different kernels to five letters: 'a', 'B', 'D', 'e', 'M'. The results are shown in Figure~\ref{fig:kernel_number}. In Figure~\ref{fig:kernel_number} the error drops as the kernel number increases from zero to $\frac{1}{10}^{th}$ of the total sample points. 
% As the kernel number increases, the error decreases. 

From the results above, we find that when we increase the kernel number to $1/10^{th}$ of the point number, the performance of the original DMP algorithm reaches high level and there is no significant benefit of adding  more kernels. Consequently, to achieve high performance with random number of sample points and to trade off the time taken by calculating the weights of too many kernels, moderate kernel number is need, like $1/10$ of the sample point number.

According to our findings, the kernels with good choice of number and shape will improve both  accuracy and efficiency. Based on our conclusions, we modified the original DMPs to a more generalized in Equations~\ref{transform_sys}-\ref{kernel_shape}, where N equals to $[Num/10]$, $Num$ is the total number of sample points:
\begin{align} \ddot{y}& =\alpha_{z}(\beta_{z}(g-y)-\dot{y})+(g-y_{0})f \label{transform_sys} \end{align} 
\begin{align}\dot{x}& =-x/T \label{linear_decay}\end{align}
% \begin{align} f& = \frac{\sum\limits_{i=1}^{N}\Psi_{x}w_{i}}{\sum\limits^{N}_{i=1}\Psi_{x}}x  \end{align}
\begin{align}\Psi_{x}& =\begin{cases} \exp(-\frac{h_{i}}{2} (x-c_{i})^{2}),& \text{if}\ x-c_{i}\leq{1/2N} \\ 0, & \text{otherwise} \end{cases} \label{kernel_shape} \end{align}

\begin{figure}[t]
\centering
\includegraphics[width=0.475\textwidth, height = 7cm]{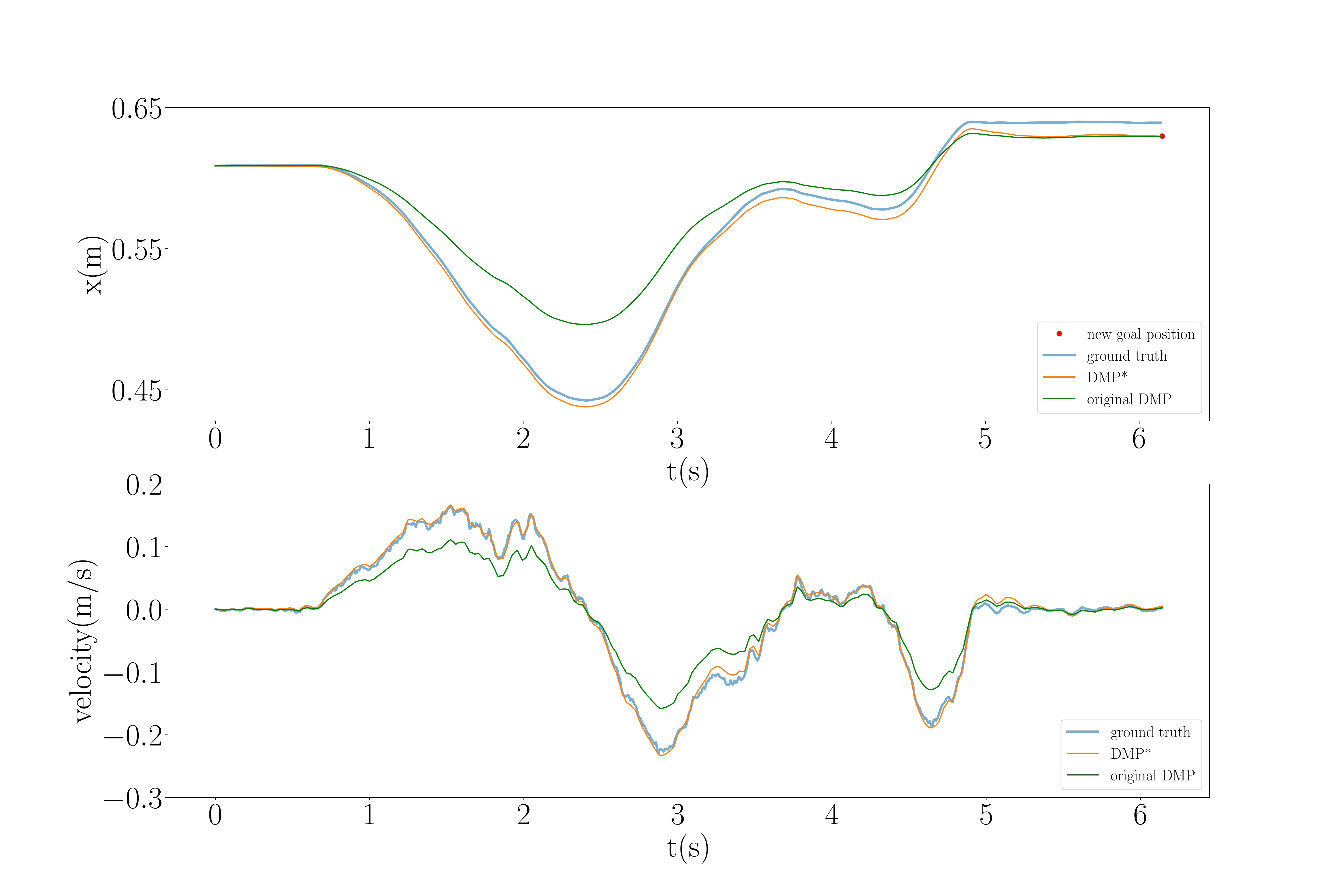}
\caption{Thie figure depicts the position and velocity calculated by the DMP* versus the original DMP in for a single DoF when changing the goal position for the intended the trajectory.}\label{fig:pos_change}
\end{figure}

\subsection{Goal-Changing in DMPs}
\label{sec:GoalChanging}
Though our newly-generalized formulation may work well in cases where start state and end state are same as demonstration, it may not be useful when such states are changing \cite{ginesi2019dmp++}. That's to say, if we fix the start point, we have to deal with the situation where the end point is changing, which can be called ``Goal-Changing Situation''. While using both original DMPs and our DMP*, some problems occur when the final goal position changes: First, if goal, $g$, is equal to the start location, $f$, the non-linear term will always be zero. Thus, the system will stay at $x_{0}$ all the time. Secondly, if $g$ is close $x_{0}$, a small change in $g$ will cause huge accelerations, which can break the limits of robot.

To solve this problem, Peter et al.~\cite{5152385}. proposed the following function in substitution of Equation~\ref{transform_sys}:
\begin{align} \ddot{y}& =\alpha_{z}(\beta_{z}(g-y)-\dot{y})-\alpha_{z}\beta_{z}(g-y_{0})x+\alpha_{z}\beta_{z}f \end{align}

\begin{figure}[b]
\centering
\includegraphics[width=0.4\textwidth,]{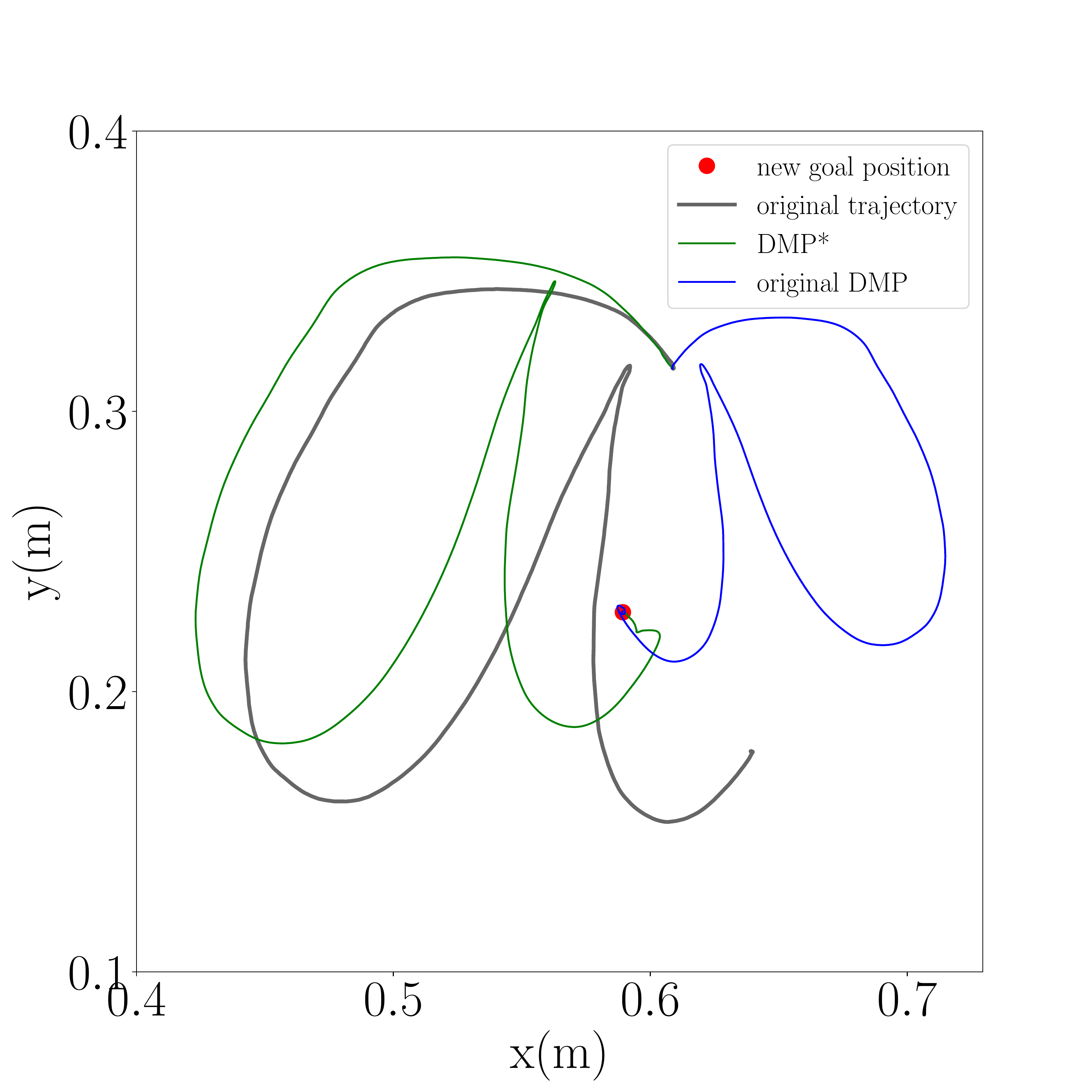}
\caption{This figure compares the performance of our DMP* algorithm versus the original DMP model when changing the goal position for writing the letter 'a'.}\label{fig:result2}
\end{figure}
\begin{figure*}[t!]
% \centering
\includegraphics[width=1.0\textwidth]{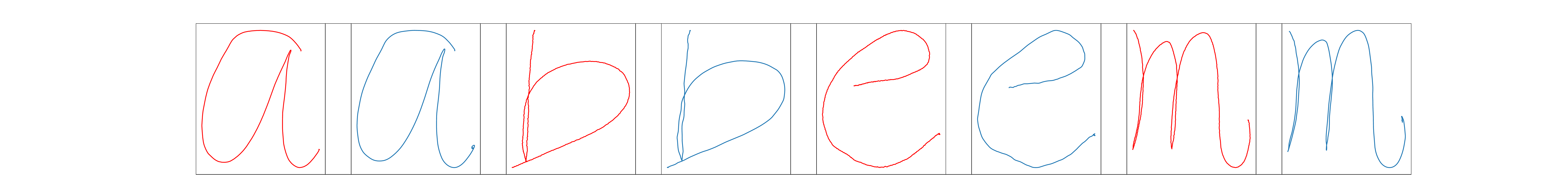}
\caption{This figure provides a comparison of training examples in red to the recreation by our algorithm, DMP*, in blue.}\label{fig:result3}
\end{figure*}

\begin{figure*}
% \centering
\includegraphics[width=1.0\textwidth]{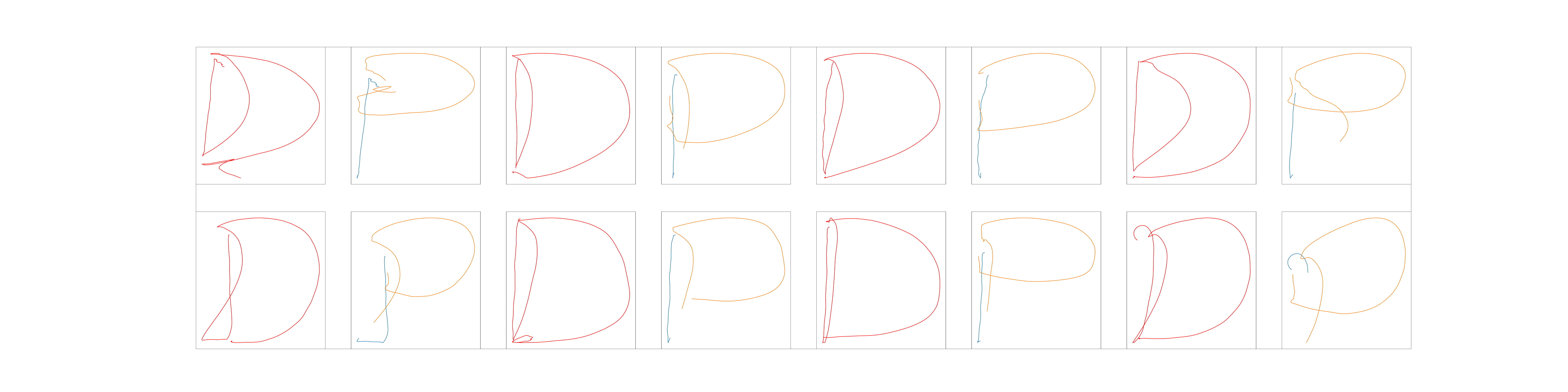}
\caption{This figure provides a performance comparison of our DMP* algorithm and the original DMP algorithm to changing the goal location for letter writing. Demonstrated examples of 'D' are shown in red. The corresponding character generated by DMP* is shown immediately to the right, consisting of the first stroke in blue and the second stroke in orange. For DMP*, we vary the goal position to show that we can change a 'D' into a 'P' seamlessly, with similar writing style. }\label{fig:result1}
\end{figure*}

Since the constant $\alpha_{z}\beta_{z}$ is used in substitution of $g-y_{0}$, the non-linear term is not influenced by the position of goals anymore. It stores the shape of original trajectory in the memory, and in new situation, it applies the learned shape to create new trajectory. Also, a decay term is added to avoid jumps at the beginning of the movement. By combining our former algorithm and Peter's et al.~\cite{5152385} modification together, we arrive at our DMP* algorithm, as shown in Algorithm \ref{dmp*_DMP}.

The performance of our DMP* exceeds the original DMP as shown in Figure~\ref{fig:pos_change} and Figure~\ref{fig:result2}. When the goal position changes 0.01 meters in Figure~\ref{fig:pos_change}, the position and velocity computed by our DMP* in 1DoF is robust to maintain the same shape as ground truth while the original DMP has relative large error. When the goal position changes 0.1 meter in 2DoF in Figure~\ref{fig:result2}, it is clearly that trajectory calculated by our DMP* still maintain highly similar shape with only changed tail at the end but the original DMP trajectory has large jumps in $x$ direction, which resulted the letter mirrored with ground truth.

% \begin{figure}[htbp]
% \centering
% \includegraphics[width=0.5\textwidth]{./pictures/xy_and_dxyz.pdf}
% \label{fig:xy_and_dxyz}
% \end{figure}

\subsection{Stroke Segmentation}
\label{sec:strokeSeg}
We employed the difference method~\cite{8694214} to perform stroke segmentation. The trajectory information given by our motion capture equipment is the 3D position of points in the form of $(x, y, z)$ sequenced by time. We invited 15 people to write on the plane and collect the moving trajectory of the sensor attached to finger. Therefore, the $z$ position of those trajectory points are equal when writing continuously. If there is a new stroke, the human will lift up the pen and there should be an acceleration, velocity and position increase along the $z$ direction. For example, when we are writing the letter 'D', there will be a lift up between we finished the first stroke and started the second stroke. 
% We calculated the velocity and acceleration as shown in Figure~\ref{fig:xy_and_dxyz} based on the position and time information.

% \begin{figure}[t]
% \centering
% \includegraphics[width=0.4\textwidth]{./pictures/robot.jpg}
% \caption{The Robot }\label{fig:robot}
% \end{figure}

The segmentation criterion according to Equation~\ref{velocity} is when the moving velocity along z direction exceeds some threshold $\theta$, the demonstrator is thought to end this stroke and to start another stroke. Equation~\ref{position} governs when the z position of the trajectory is far away from the writing plane. By combining Equations~\ref{velocity}-\ref{position}, we can determine whether the stroke has finished. Empirically, we find that the difference method~\cite{8694214} augmented for stroke segmentation in handwriting tasks works efficiently on our data set.
\begin{align} \frac{dz}{dt} > \theta \label{velocity}\end{align}
\begin{align} |z-\mu_z| < \sigma  \label{position} \end{align}

% However, there are also some trajectories that can not be segmented perfectly by only using the heuristic functions. We still see some extra parts on the plane which is the part when the demonstrator lifting up its hand with slow speed and close to the writing plane. In the future, we will consider combining this method with Hidden Markov Model (HMM) or Gaussian Mixture Model (GMM) to achieve a more generalized segmentation.

\section{EXPERIMENTAL EVALUATION}
\label{sec:evaluate}
In this section, we introduce how we built the handwriting system, and show the performance of this system.

\subsection{System Architecture}
As shown in Figure~\ref{fig:pipeline}, our robot handwriting learning system is consisted with a motion capture system, a master computer for processing the collected data and sending out the processed data to the Fetch robot through ROS, marker pens for writing and A4 size papers. The platform we use to collect data is a motion capture equipment functioned by laser emitters and sensors with accuracy achieving to at least $10^{-3}$ meter. The equipment can provide us with ‘target trajectory’ from human teacher. We test on DMP* algorithm in different situations and for different writing styles.
% \subsection{Analysis Robustness of DMP++}

\subsection{Letter generation based on DMP* }
% \subsubsection{show the result of replicated letters}
% \subsubsection{show the result of replicated and created letters}

Firstly we apply our DMP* algorithm to letter recreation based on different letters. In Figure~\ref{fig:result3}, we collect human handwriting trajectories of four letters: 'a', 'b', 'e', 'm.' which are shown in red. By using DMP*, we successfully reproduce the letters in nearly identical shapes, which are shown in blue trajectories in Figure \ref{fig:result3}. 

%\mcgnote{NEED TO RE-WRITE TO SAY THAT YOU ARE GOING TO 1) DRAW MULTI-STOKE CHARACTERS AND 2) DRAW MULTI-STROKE CHARACTERS BY COMPOSING SKILLS FROM SIMPLER TASKS.}

Having shown the accuracy of our DMP* in reproducing a single-stroke letter, such as 'a', we now apply DMP* to more complicated, multi-stroke letters, which we generate by combining sub-skills to perform a more complex writing task. Due to the robustness of the DMP* algorithm, the robot can change the ending position as it desired and still preserve the original writing style. Then the robot can combine two changed strokes and create a new letter with similar strokes learned from previous letter. Obviously, the newly created letter has has similar writing style as the demonstration letter.

In our experiment, we segment letter 'D' into two strokes, one of which is a straight line and the other is a curve line. Then we apply DMP* to learn these two trajectories and create letter 'P' of similar writing style through two strokes. The results are shown in the Figure~\ref{fig:result1}. Lastly, we use the Fetch robot to reproduce the learned trajectory, which is shown in the first image Figure~\ref{fig:robot_writing} in our paper.

\section{DISCUSSION}
%\mcgnote{QIAN -- HERE, RE-SUMMARIZE YOUR CONTRIBUTIONS. 1 PARAGRAPH IS GOOD.}
The results shown in Section \ref{sec:evaluate} prove the high accuracy and the robustness of our robotic handwriting system. Our handwriting-learning system provides a generalized and practical way to learn trajectories of arbitrary shape, arbitrary scale, arbitrary strokes and with arbitrary starting and ending positions. From the demonstrations, we can not only reproduce handwriting trajectories of identical shape, but also transfer the writing style to create other trajectories. Also, our proposed DMP* serves as a more accurate method to imitate trajectories, comparing to state-of-art DMPs, and could be applied as a general way to solve learning from demonstration problems.

\section{CONCLUSION}
\label{sec:conclusions}
In this paper, we proposed a generalized handwriting learning system from demonstrator data collection to trajectory reproduction. Experimental results have shown that our DMP* algorithm outperforms original DMPs algorithm with better accuracy and style transfer ability. These identities enable our robot to achieve multiple tasks with few shots learning, making it more applicable in the real world application. 

In the future, we will test the performance of more complex trajectories on our robotic handwriting-learning system. Also, we would further improve our DMP* algorithm and apply it to more learning from demonstration scenes.

% \section{FUTURE WORK}
% \label{sec:futurework}

% \addtolength{\textheight}{-12cm}    % This command serves to balance the column lengths
                                  % on the last page of the document manually. It shortens
                                  % the textheight of the last page by a suitable amount.
                                  % This command does not take effect until the next page
                                  % so it should come on the page before the last. Make
                                  % sure that you do not shorten the textheight too much.

%%%%%%%%%%%%%%%%%%%%%%%%%%%%%%%%%%%%%%%%%%%%%%%%%%%%%%%%%%%%%%%%%%%%%%%%%%%%%%%%

%%%%%%%%%%%%%%%%%%%%%%%%%%%%%%%%%%%%%%%%%%%%%%%%%%%%%%%%%%%%%%%%%%%%%%%%%%%%%%%%

%%%%%%%%%%%%%%%%%%%%%%%%%%%%%%%%%%%%%%%%%%%%%%%%%%%%%%%%%%%%%%%%%%%%%%%%%%%%%%%%
% \section*{APPENDIX}

% Appendixes should appear before the acknowledgment.

%%%%%%%%%%%%%%%%%%%%%%%%%%%%%%%%%%%%%%%%%%%%%%%%%%%%%%%%%%%%%%%%%%%%%%%%%%%%%%%%

\bibliography{ref}

\begin{thebibliography}{10}

\bibitem{ginesi2019dmp++}
Michele Ginesi, Nicola Sansonetto, and Paolo Fiorini.
\newblock Dmp++: Overcoming some drawbacks of dynamic movement primitives.
\newblock {\em arXiv preprint arXiv:1908.10608}, 2019.

\bibitem{childrenteach}
D.~{Hood}, S.~{Lemaignan}, and P.~{Dillenbourg}.
\newblock When children teach a robot to write: An autonomous teachable
  humanoid which uses simulated handwriting.
\newblock In {\em 2015 10th ACM/IEEE International Conference on Human-Robot
  Interaction (HRI)}, pages 83--90, March 2015.

\bibitem{ijspeert2002learning}
Auke~Jan Ijspeert, Jun Nakanishi, and Stefan Schaal.
\newblock Learning attractor landscapes for learning motor primitives.
\newblock In {\em Proceedings of the 15th International Conference on Neural
  Information Processing Systems}, pages 1547--1554. MIT Press, 2002.

\bibitem{buildingsuc}
A.~{Jacq}, S.~{Lemaignan}, F.~{Garcia}, P.~{Dillenbourg}, and A.~{Paiva}.
\newblock Building successful long child-robot interactions in a learning
  context.
\newblock In {\em 2016 11th ACM/IEEE International Conference on Human-Robot
  Interaction (HRI)}, pages 239--246, March 2016.

\bibitem{johal2016child}
Wafa Johal, Alexis Jacq, Ana Paiva, and Pierre Dillenbourg.
\newblock Child-robot spatial arrangement in a learning by teaching activity.
\newblock In {\em 2016 25th IEEE International Symposium on Robot and Human
  Interactive Communication (RO-MAN)}, pages 533--538. Ieee, 2016.

\bibitem{khansari2011learning}
S~Mohammad Khansari-Zadeh and Aude Billard.
\newblock Learning stable nonlinear dynamical systems with gaussian mixture
  models.
\newblock {\em IEEE Transactions on Robotics}, 27(5):943--957, 2011.

\bibitem{kober2011reinforcement}
Jens Kober, Erhan Oztop, and Jan Peters.
\newblock Reinforcement learning to adjust robot movements to new situations.
\newblock In {\em Twenty-Second International Joint Conference on Artificial
  Intelligence}, 2011.

\bibitem{6008670}
T.~{Kulvicius}, K.~{Ning}, M.~{Tamosiunaite}, and F.~{Worgötter}.
\newblock Joining movement sequences: Modified dynamic movement primitives for
  robotics applications exemplified on handwriting.
\newblock {\em IEEE Transactions on Robotics}, 28(1):145--157, Feb 2012.

\bibitem{kulvicius2011modified}
Tomas Kulvicius, KeJun Ning, Minija Tamosiunaite, and Florentin
  W{\"o}rg{\"o}tter.
\newblock Modified dynamic movement primitives for joining movement sequences.
\newblock In {\em 2011 IEEE International Conference on Robotics and
  Automation}, pages 2275--2280. IEEE, 2011.

\bibitem{2012levine}
Sergey Levine and Vladlen Koltun.
\newblock Continuous inverse optimal control with locally optimal examples.
\newblock In {\em ICML '12: Proceedings of the 29th International Conference on
  Machine Learning}, 2012.

\bibitem{8694214}
C.~{Li}, C.~{Yang}, and C.~{Giannetti}.
\newblock Segmentation and generalisation for writing skills transfer from
  humans to robots.
\newblock {\em Cognitive Computation and Systems}, 1(1):20--25, 2019.

\bibitem{muelling2010learning}
Katharina Muelling, Jens Kober, and Jan Peters.
\newblock Learning table tennis with a mixture of motor primitives.
\newblock In {\em 2010 10th IEEE-RAS International Conference on Humanoid
  Robots}, pages 411--416. IEEE, 2010.

\bibitem{5152385}
P.~{Pastor}, H.~{Hoffmann}, T.~{Asfour}, and S.~{Schaal}.
\newblock Learning and generalization of motor skills by learning from
  demonstration.
\newblock In {\em 2009 IEEE International Conference on Robotics and
  Automation}, pages 763--768, May 2009.

\bibitem{potkonjak2012robot}
Veljko Potkonjak.
\newblock Robot handwriting: Why and how?
\newblock In {\em Interdisciplinary Applications of Kinematics}, pages 19--35.
  Springer, 2012.

\bibitem{rohrbeck2003peer}
Cynthia~A Rohrbeck, Marika~D Ginsburg-Block, John~W Fantuzzo, and Traci~R
  Miller.
\newblock Peer-assisted learning interventions with elementary school students:
  A meta-analytic review.
\newblock {\em Journal of educational Psychology}, 95(2):240, 2003.

\bibitem{7759554}
{Ruohan Wang}, Y.~{Wu}, {Wei Liang Chan}, and {Keng Peng Tee}.
\newblock Dynamic movement primitives plus: For enhanced reproduction quality
  and efficient trajectory modification using truncated kernels and local
  biases.
\newblock In {\em 2016 IEEE/RSJ International Conference on Intelligent Robots
  and Systems (IROS)}, pages 3765--3771, Oct 2016.

\bibitem{SCHAAL2007425}
Stefan Schaal, Peyman Mohajerian, and Auke Ijspeert.
\newblock Dynamics systems vs. optimal control — a unifying view.
\newblock In Paul Cisek, Trevor Drew, and John~F. Kalaska, editors, {\em
  Computational Neuroscience: Theoretical Insights into Brain Function}, volume
  165 of {\em Progress in Brain Research}, pages 425 -- 445. Elsevier, 2007.

\bibitem{ude2010task}
Ale{\v{s}} Ude, Andrej Gams, Tamim Asfour, and Jun Morimoto.
\newblock Task-specific generalization of discrete and periodic dynamic
  movement primitives.
\newblock {\em IEEE Transactions on Robotics}, 26(5):800--815, 2010.

\bibitem{williams2006extracting}
Ben~H Williams, Marc Toussaint, and Amos~J Storkey.
\newblock Extracting motion primitives from natural handwriting data.
\newblock In {\em International Conference on Artificial Neural Networks},
  pages 634--643. Springer, 2006.

\bibitem{Wu2018MultiModalRA}
Yun Wu, Ruohan Wang, Luis~Fernando D'Haro, Rafael~E. Banchs, and Keng~Peng Tee.
\newblock Multi-modal robot apprenticeship: Imitation learning using linearly
  decayed dmp+ in a human-robot dialogue system.
\newblock {\em 2018 IEEE/RSJ International Conference on Intelligent Robots and
  Systems (IROS)}, pages 1--7, 2018.

\bibitem{yin2016synthesizing}
Hang Yin, Patr{\'\i}cia Alves-Oliveira, Francisco~S Melo, Aude Billard, and Ana
  Paiva.
\newblock Synthesizing robotic handwriting motion by learning from human
  demonstrations.
\newblock In {\em Proceedings of the 25th International Joint Conference on
  Artificial Intelligence}, number CONF, 2016.

\end{thebibliography}
\bibliographystyle{plain}
\end{document}